\documentclass{article}

\usepackage{arxiv}

\usepackage[utf8]{inputenc} 
\usepackage[T1]{fontenc}    
\usepackage{hyperref}       
\usepackage{url}            
\usepackage{booktabs}       
\usepackage{nicefrac}       
\usepackage{microtype}      
\usepackage{lipsum}		
\usepackage{natbib}
\usepackage{doi}
\setcitestyle{numbers,square}
\usepackage{amsmath,amssymb,amsfonts}
\usepackage{graphicx,color}
\usepackage{textcomp}
\usepackage{xcolor}
\usepackage{algorithm,algorithmic}
\usepackage{threeparttable} 
\usepackage{multirow}
\usepackage{subcaption} 

\title{Automatic Assessment of Students' Classroom Engagement with Bias Mitigated Multi-task Model\thanks{{This paper is currently under review at a publisher.}}}

\author{ {\hspace{1mm}James Thiering} \\
	Computer Science \& Engineering,\\
	University of Colorado Denver,\\
	Colorado, CO 80204, \\
	\texttt{james.thiering@ucdenver.edu} \\
    \And
    {\hspace{1mm}Tarun Sethupat Radha Krishna} \\
	Computer Science \& Engineering,\\
	University of Colorado Denver,\\
	Colorado, CO 80204, \\
	\texttt{tarun.sethupatradhakrishna@ucdenver.edu} \\
    \And
    {\hspace{1mm}Dylan Zelkin} \\
	Computer Science \& Engineering,\\
	University of Colorado Denver,\\
	Colorado, CO 80204, \\
	\texttt{dylan.zelkin@ucdenver.edu} \\
	\And
	\href{https://orcid.org/0000-0002-7446-4639}{\includegraphics[scale=0.06]{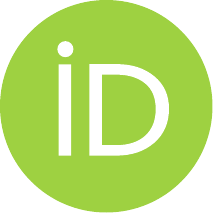}\hspace{1mm}Ashis Kumer Biswas} \\
	Computer Science \& Engineering,\\
	University of Colorado Denver,\\
	Colorado, CO 80234, \\
	\texttt{ashis.biswas@ucdenver.edu}
}

\date{October 24, 2025}

\renewcommand{\shorttitle}{Automatic Assessment of Students' Classroom Engagement}

\hypersetup{
pdftitle={Automatic Assessment of Students' Classroom Engagement with Bias Mitigated Multi-task Model},
pdfsubject={AI},
pdfauthor={James Thiering,Tarun Sethupat Radha Krishna,Dylan Zelkin, Ashis Kumer Biswas},
pdfkeywords={Attribute orthogonal regularization, multi-task learning, e-learning, engagement}
}

\begin{document}
\maketitle

\begin{abstract}
With the rise of online and virtual learning, monitoring and enhancing student engagement have become an important aspect of effective education. Traditional methods of assessing a student’s involvement might not be applicable directly to virtual environments. In this study, we focused on this problem and addressed the need to develop an automated system to detect student engagement levels during online learning. We proposed a novel training method which can discourage a model from leveraging sensitive features like gender for its predictions. The proposed method offers benefits not only in the enforcement of ethical standards, but also to enhance interpretability of the model predictions. We applied an attribute-orthogonal regularization technique to a split-model classifier, which uses multiple transfer learning strategies to achieve effective results in reducing disparity in the distribution of prediction for sensitivity groups from a Pearson correlation coefficient of 0.897 for the unmitigated model, to 0.999 for the mitigated model. The source code for this project is available on \url{https://github.com/ashiskb/elearning-engagement-study}.
\end{abstract}


\section{Introduction}
Bias mitigation in machine learning is important due to the introduction of such models in everyday life. There are devices and systems in place that intake data and parse it through black-box systems that we hope produce acceptable results. To understand why a model produces the results that it does, a data scientist must make certain assumptions about the way in which the model was trained, and the data that was used for training. One assumption is that there is no unaddressed bias in the dataset that the model may leverage. Unfortunately, all datasets will contain some bias \cite{bird2020fairlearn}, even when a dataset contains  an equal number of samples for all groups \cite{wang2019balanced}.

The main issue of bias in machine learning models is spurious correlation, which occurs when two variables are correlated, but do not have a causal relationship \cite{haig2003spurious}. In the case of vision tasks, features that a model uses for prediction can be categorized as either non-spurious core features or spurious features \cite{singla2021salient,teotia2022finding}. It is common for deep learning tasks to use spurious features for predictions \cite{teotia2022finding,arjovsky2019invariant}. An issue also stems from suppressing spurious features, which can degrade model performance \cite{khani2021removing}. In order to address these issues, there are different approaches to model design that can affect the resulting bias \cite{yang2022enhancing,lou2013accurate}. Our approach in this research is to address the bias through model design and training methods, rather than manipulating the dataset. Through implementation of specific regularization techniques as well as multiple-transfer learning, we showed that a model can be designed to be intrinsically debiased. 

As a case study, this paper details the exploratory data analysis, training, and evaluation of models for an engagement grading task. An engagement classification model has the potential to facilitate improved online classroom learning experiences through feedback for teachers. Due to the unique and specific nature of this task, there is limited labelled data available. The DAiSEE dataset \cite{gupta2016daisee} has labels for engagement, and will be the focus of this research.

\textbf{Problem Statement}: There are several issues with the DAiSEE dataset. The distribution of  ground truth labels for engagement level is skewed such that there are a disproportionate number of high engagement level samples. Additionally, and most importantly, there is a skew in the distribution of levels between genders of participants such that females were, on average, annotated as being more engaged than the male participants. Due to the skew in label distribution in the DAiSEE dataset, models trained on this data display a gender bias in predictions, and mitigating techniques are necessary to reduce the bias present in the model.

\textbf{Social Impact of Bias}: Machine learning systems have the potential to cause fairness-related harm \cite{bird2020fairlearn}, so it is the responsibility of ethical AI practitioners to evaluate the impact that systems have on demographic groups. For the use case of an engagement classifier for online classrooms, a model with bias toward certain demographic groups may lead to either an over-estimation of class engagement or under-estimation of class engagement. Both scenarios could have significant ramifications given long-term exposure to feedback from such a system. Teachers may be more frustrated in classes with demographics which bias the model toward low engagement. Likewise, teachers may not take necessary interventions to improve class engagement when demographics bias the model predictions toward high engagement, leading to underperformance of certain demographic groups. Cumulative effects of such a system can have compounding impact \cite{bird2020fairlearn}.

Training a model on a desired task is fundamental to the value of the predictions made by a deep learning model. If a model leverages spurious features to make predictions, it is an issue that can cause out-of-distribution predictions to be unreliable \cite{teotia2022finding}. Because of this, mitigation techniques can be useful outside of classification in human-related tasks. Examples of bias in image datasets not pertaining to protected demographic groups can include environmental conditions, lighting, viewpoint, and background objects \cite{teotia2022finding,arjovsky2019invariant,khani2021removing}. 

Our research can be described as accomplishing two major tasks. The first accomplishment is the detection of bias in the DAiSEE dataset through exploratory data analysis. The analysis uncovered the skew in the distribution of labels overall, as well as the skew in distribution of labels between genders. The second accomplishment is the proposal of a solution to mitigate the bias present in the dataset. A novel training regimen is described which reduces the bias in predictions between genders, and additional research is outlined for potential improvements.

This paper will first review some of the principal concepts that are important to the project in section 2 This includes concepts such as transfer learning, Generalized Additive Models, mitigation techniques such as sample reweighting and adversarial training, as well as findings from research in multiple multi-task training architectures.

Section 2 describes the DAiSEE dataset in detail. This section of the paper describes the exploratory data analysis performed and the findings of skewed distributions. Additionally, this section discusses the ambiguity of the origin of the bias and goes into detail on some missing details that are critical for developing a model using the dataset.

Section 4 discusses our proposed methods for training a model using mitigating techniques. The model architecture and training setup is described for the purpose of reproducibility. The loss functions, regularization term, and steps taken to train the model are described. 

Section 5 analyzes the results of training in three primary areas: bias in the distribution of predictions, F1 score performance, and accuracy performance. Interpretation of these results is highlighted as a significant challenge.

Section 6 concludes the paper by summarizing the findings and proposing new research. The research proposed will help improve the bias mitigation techniques described in this paper as well as help improve the DAiSEE dataset with an additional validation study.

\section{Related Works}
\subsection{Transfer Learning in Deep CNN Models}
Pre-trained deep convolutional neural networks (DCNN) models are commonly used as feature extractors in transfer learning applications, and are widely used for tasks such as facial emotion/expression recognition (FER) \cite{akhand2021facial}. Transfer learning is particularly beneficial for visual models because of the early layer convolution filters that are common across most tasks. The sharing of these layers is accomplished by training a model on a large set of natural images such as ImageNet \cite{deng2009imagenet}. The pre-trained model is referred to as a source model or base learner. The convolutional layers of this model are then preserved, while subsequent fully connected (also known as dense) layers that are associated with final layers are discarded. The weights associated with the early layers are then frozen and used as a feature extractor for a target task. This transfer improves the performance of target task models which have a limited number of samples \cite{pan2009survey}. Furthermore, it decreases the time it takes to train and develop the model by providing a framework that has been demonstrated to perform well in natural image tasks.

Common models such as Inception, Xception, VGG, ResNet, as well as others, are available with pretrained weights through libraries such as Keras. Once imported, the model structure is then adapted to the desired task and trained with the pretrained layer weights locked. After the classifier layers are trained, further training can be performed throughout the entire model with a reduced learning rate \cite{agrawal2014analyzing}. This process of additional training is referred to as fine-tuning and can have significant impact on model performance.

\subsection{Generalized Additive Models}
The type of data available to the model largely determines the approaches to bias mitigation. In situations when samples can be described by a feature vector of discrete and independent measurements, generalized additive models (GAMs) are a tractable solution to bias mitigation. It was found that GAMs with pair-wise interactions can perform nearly as well as simple neural networks, and still retain intelligibility \cite{lou2013accurate}.

In the case of computer vision tasks, the problem of explainability is exacerbated by the spatial relationship of the features in the image. Vision models cannot be decomposed into univariate models. However, if a feature extractor can have output values individually corelated with certain explainable features, this could be a path toward decomposing the output of a feature extractor into univariate models. Future work will be needed to determine if this is a viable strategy for bias mitigation.

\subsection{Sample Reweight}
Bias mitigation techniques involving sample weighting have been demonstrated to help but have limitations \cite{yang2022enhancing}. Sample weighting to address sensitivity group imbalances requires that the training samples have labels for each sensitivity group, which can add a large amount of work and cost associated with dataset preparation. Additionally, it has been shown to have limited effect \cite{yang2022enhancing}. More importantly, it does not address underlying bias in the existing target labels.

\subsection{Adversarial Models}
\begin{figure}[!h]
    \centering
    \includegraphics[scale=01.0]{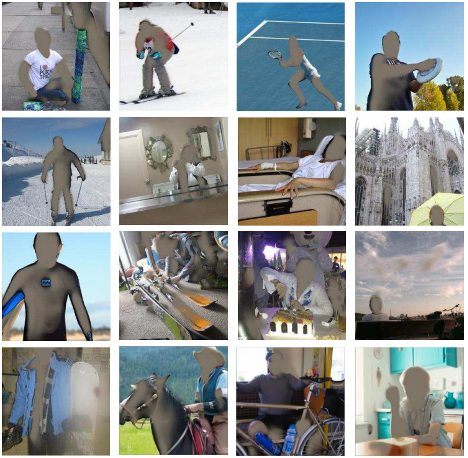}
    \caption{Adversarial-trained feature suppression from \cite{wang2019balanced}}
    \label{fig:adversarial-trained-feat-suppression}
\end{figure}

Adversarial models have been used to remove features associated with gender as a form of bias mitigation \cite{wang2019balanced,yang2022enhancing}. In work performed in \cite{wang2019balanced}, it was noted that adversarial removal of gendered features could remove the entire person or entire face (see Fig. \ref{fig:adversarial-trained-feat-suppression}). Because of this, adversarial removal may not be appropriate for tasks which require visual observation of features, such as faces, which are related to the sensitivity group in question.  It is also noted that spurious feature suppression through adversarial training can also disproportionately affect the accuracy between sensitivity groups \cite{khani2021removing,yang2022enhancing}. 

\subsection{Split-Model Design}
For multi-task training, there are different structures that can be used to obtain varying results. In \cite{madan2020and}, the task is to train a model that classifies cars by type and to predict a spatial component related to the orientation around the car. 
The three model architectures that were explored are shared models, where the classifier layers bifurcate from the main branch immediately prior to classification, a separate model where the two classifier branches bifurcate immediately after the input, and a split model, where the classifier branches bifurcate after several shared feature extractor layers \cite{madan2020and}. It was found that the separate branches performed best due to increased specialization. However, the split-2 model achieved a performance score between the two. It is hypothesized that specialization of neurons allows for greater invariance of representation .


\section{The dataset and associated challenges}
As a case study, we have used the DAiSEE dataset \cite{gupta2016daisee} as our focus of research. This dataset is unique because it contains labels for engagement levels. Similar research has taken place on engagement classification, but datasets were not made available publicly \cite{mohamad2020automatic}. Because DAiSEE has labels for engagement, we were able to use it to train models though supervised learning.

The DAiSEE dataset \cite{gupta2016daisee} is comprised of 9,068 video files containing 10 seconds of “in the wild” video recording. Each file has associated labels for four affective states: boredom, confusion, engagement, and frustration. Each affect label has an integer representing affect level, varying from zero (very low) to three (very high). Because of this, each sample has a four-by-four matrix for ground truth. In addition to the affective states label, there is also a binary label for gender ground truth.

Each sample was annotated using crowd annotations through a service provided by CrowdFlower, now Appen Limited, which uses a validation technique between annotators to ensure consistency between annotators. Each video snippet received label votes from 10 different annotators \cite{gupta2016daisee}. This labelling style differs from other datasets collected for similar tasks such as \cite{mohamad2020automatic}, where psychology students were the annotators for the entire set.

\begin{figure}[t]
    \centering
    \includegraphics[scale=0.5]{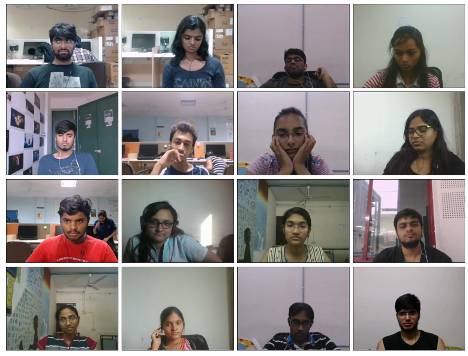}
    \caption{Samples of Different Engagement Levels. Top row: very-low engagement; Second row: low engagement; Third row: high engagement; Bottom row: very-high engagement.}
    \label{fig:samples-engagement-levels}
\end{figure}

Samples are extracted from the video clips to produce an image dataset. Examples of these images are shown in Fig. \ref{fig:samples-engagement-levels}, where random samples are drawn from the four represented levels: the top row shows examples of very low engagement, the second row shows low engagement, the third row shows high engagement, and the bottom row shows very-high engagement.

The decision to annotate the labels with four levels was to discourage a middle “neutral” level. Additionally, an even number of levels allows the labels to be converted into a binary classification problem for high and low engagement scores \cite{gupta2016daisee}. The choice to annotate with levels instead of binary labels for affective states differs from other datasets relating to facial expression recognition, which tend to have discrete affective states for classification tasks \cite{lundqvist1998karolinska,goeleven2008karolinska}.

\subsection{Temporal Consideration}
Because the samples are labeled once throughout a ten second duration, the individually extracted frames often do not represent the overall nature of the sample over the entire time segment. This is because the display of engagement can vary throughout the short time-span of the sample \cite{mohamad2020automatic}. 

\begin{figure}
    \centering
    \includegraphics[scale=0.5]{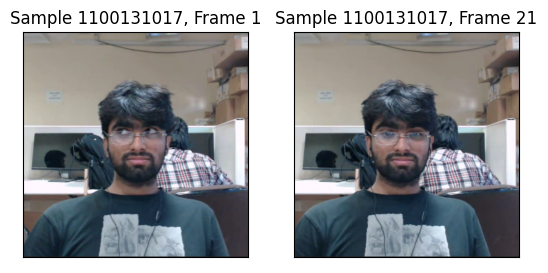}
    \caption{Sample frames displaying contradicting engagement levels}
    \label{fig:samples-contradicting-engagement-levels}
\end{figure}

In Fig \ref{fig:samples-contradicting-engagement-levels}, a DAiSEE sample (id: 1100131017) exemplifies this problem. The sample has a label of very low engagement. The first frame appears representative of this label, however frame 21 in isolation does not match this ground truth label. The full 10 second video clip that the frames were extracted from appears to have the correct ground truth level of very low engagement because the participant spends a large portion of the time looking away from the screen.

Temporal considerations for engagement classification have been demonstrated to be effective at improving model accuracy. In \cite{yang2022enhancing}, a residual network (resnet) is used as a feature extractor for individual frames. The output of the resnet feature extractor is then concatenated and analyzed by a temporal convolutional network (TCN) for engagement classification. Additionally, a resnet feature extractor was combined with a long short-term memory (LSTM) recurrent neural network (RNN) for comparison. The resnet/TCN model achieved an accuracy of 63.9\%, while the resnet/LSTM model performed with an accuracy of 61.15\%. The temporal nature of the affective states is recognized by \cite{gupta2016daisee,mohamad2020automatic,abedi2021improving}.

\subsection{Label Skew}
\begin{figure}[!h]
    \centering
    \includegraphics[scale=0.75]{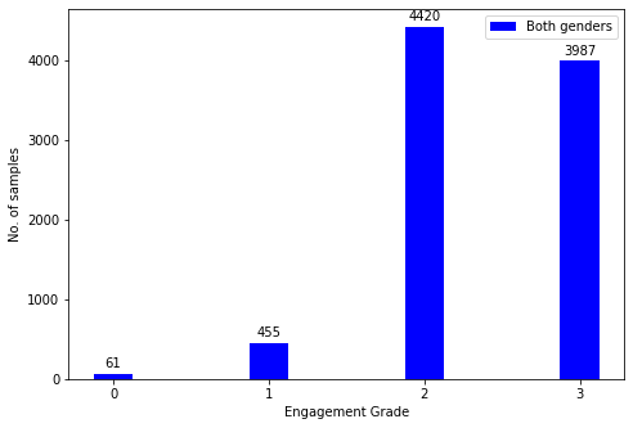}
    \caption{Distribution of labels}
    \label{fig:distribution-of-labels}
\end{figure}

\begin{figure}[!h]
    \centering
    \includegraphics[scale=0.75]{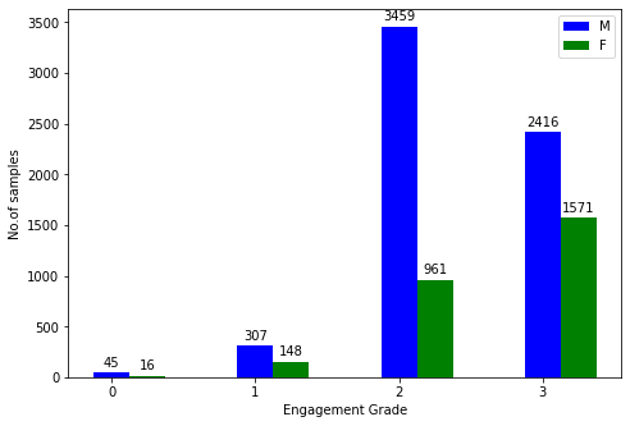}
    \caption{Distribution of labels, separated by gender}
    \label{fig:distribution-of-labels-gender}
\end{figure}

Two main issues exist regarding the distribution of the ground true levels for the engagement labels. The overall distribution of labels is skewed toward ground truth levels two and three (Fig. \ref{fig:distribution-of-labels}). Levels two and three represent 94\% of the entire dataset, whereas levels zero and one represent only 6\%. The mode of the label distribution is level two, which represents 49.5\% of the samples. This problem is compounded by the fact that the distribution is skewed by a gendered bias (Fig. \ref{fig:distribution-of-labels-gender}). The skew in labels between genders can be represented by a Pearson correlation coefficient of 0.82 between the two distributions. The mode of the male labels is level two, or “high engagement”. The mode of the female labels is level three, or “very-high engagement”. Additionally, there is a third important imbalance that exists in the overall male to female ratio of 2.3:1.

\subsection{Bias and Uncertainty in Ground Truth}
When bias in ground truth labels exists between sensitivity groups, it exposes an uncomfortable dilemma. The first scenario is that human bias has judged the individuals differently based on their sensitivity group. The second scenario is that females are more attentive and engaged, and the skewed distribution of the sample set is more representative of reality. It may be impossible to distinguish between these two cases, and it could be a combination of the two factors. Therefore, an ideal classifier will always produce the same distribution of predictions regardless of demographic sensitivity group.

\subsection{Human Validation}
Other datasets such as the Karolinska Directed Emotional Faces (KDEF) received a validation study to determine whether the samples were interpretable by human participants \cite{goeleven2008karolinska}. This study on the KDEF dataset resulted in a mean success index of 71.87\%, with a standard deviation of 25.78. This suggests that interpretability of facial expression/emotion recognition tasks is difficult even for human observers. Unfortunately, the DAiSEE dataset does not appear to have a validation survey completed, so it is uncertain whether the samples are interpretable by human observers, and therefore it is difficult to establish what metrics would be considered acceptable for a model. As in the case of the KDEF validation study, a subset validation study of the DAiSEE dataset would create a baseline expectation for DNN model performance. This will be further described under the Further Research section.

\section{Proposed Methodology}
\subsection{Model Architecture}
The Xception model \cite{fran2017deep} is used as the source model and feature extractor for the models are examined in this paper. A baseline model was trained using Xception as a feature extractor followed by two dense layers and a classification layer. Models that were trained in this research follow similar design patterns to this structure to reduce the number of additional factors that could influence the results. Input to the models is preprocessed using the Keras API though the preprocessing pipeline methods, and the input was constrained to image dimensions of 299 pixels in width and 244 pixels in height. Each sample is color with three channels for red, green, and blue. The preprocessing pipeline scales each input pixel to values between -1 and 1.

Architectural choices for multi-task learning can influence the ability for separate tasks to influence shared learning \cite{madan2020and}. The findings in \cite{madan2020and} report that in the case of multi-task learning, synergistic tasks may benefit from shared layers, while non-synergistic tasks may reduce performance. It is suggested that this is a result of selectivity and invariance represented by individual neurons. In the case of synergistic tasks, shared architecture can be beneficial through shared learning. In the case of non-synergistic tasks, separate architecture improves performance. It was also demonstrated that a split architecture can perform nearly as well in non-synergistic tasks as separate models.

\subsection{Grad-CAM}
\begin{figure}
    \centering
    \includegraphics[scale=0.5]{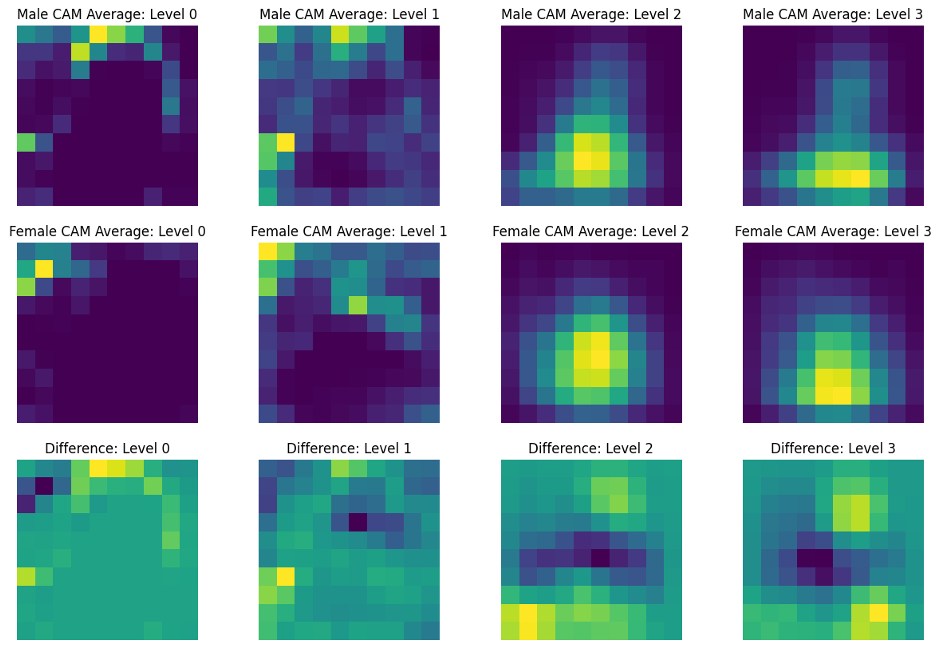}
    \caption{Grad-CAM heat map averages}
    \label{fig:grad-cam-heat-map-averages}
\end{figure}

In order to determine the bias associated with a model trained on the DAiSEE dataset, a class activation mapping algorithm (Grad-CAM) \cite{selvaraju2017grad} was used to determine the localized responses between aggregates of predictions of the two genders (Fig. \ref{fig:grad-cam-heat-map-averages}). We found that in aggregate, the model used different facial areas of the images between genders to make predictions. Unfortunately, due to the model architecture, only a low-resolution approximation of the activation mapping is available. Nonetheless, these images show that between grades levels two and three, the classifier is looking at slightly different regions of the image, which could indicate a preference for characteristics such as hair styles, body posture, sexual dimorphic features, or some unknown spurious feature. For levels zero and one, it is most likely spurious background features recognized due to the limited number of training examples in these levels.

\subsection{Attribute-Orthogonal Regularization}
An article \cite{yang2022enhancing} discusses an approach to bias mitigation in visual tasks. The article suggests penalizing the correlation between the weights of two branching classifiers from a shared feature extractor. This technique was shown to have some success at reducing bias, and was adapted for use with the engagement task. In our case, one classifier branch would output the engagement grade level prediction (cls1), while the other classifier is trained to predict gender (cls2). In effort to improve the effectiveness of this regularization technique, the multi-task training scenario has been split into multiple phases and the loss function has been augmented accordingly.

The attribute-orthogonal regularization (AOR) term $L_\text{ortho}$ represents the normalized correlation of the two layers directly following the last feature extractor layer.

\begin{eqnarray}
L_\text{ortho}=\dfrac{||\theta_\text{cls1a}\cdot \theta_\text{cls2a} ||_1}{||\theta_\text{cls1a}||_2  \cdot ||\theta_\text{cls2a}||_2 }
\end{eqnarray}

This regularization term is added to the existing categorical cross-entropy loss function of the engagement classifier.

\begin{eqnarray}
\underset{\theta_\text{cls1}}{min}\quad L_\text{cls1} + \lambda L_\text{ortho}
\end{eqnarray}
where, $L_\text{ortho} = - \sum_{i=1}^c y_i \log \hat{y}_i$

\begin{figure}
    \centering
    \includegraphics[scale=0.75]{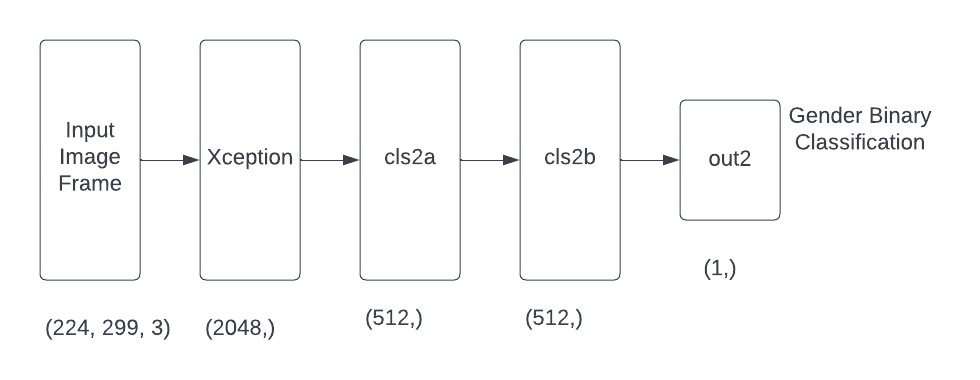}
    \caption{Gender Classifier Structure }
    \label{fig:gender-classifier-structure}
\end{figure}

These two classifiers cls1 and cls2 may be trained in stages. For example, the gender classification task may be trained initially as a separate task (Fig. \ref{fig:gender-classifier-structure}). It was found that a gender classifier trained in this way utilizing the DAiSEE dataset would only perform with an accuracy of approximately 60\% on the validation set. As this weak classifier did not significantly mitigate bias through the AOR augmented loss, it was found that instead of training a gender classifier using the DAiSEE samples, another dataset \cite{eidinger2014age} that was curated partially for gender classification was able to train a more accurate gender classifier. 

In this multi-task transfer learning scenario, the gender classifier is trained first on the separate dataset containing many in-the-wild samples with only gender labels. The dataset used was collected and curated by the Open University of Israel (OUI) and analyzed in \cite{eidinger2014age}. This dataset also has labels for age, which could be useful in additional bias mitigation, however only the gender label is used from this dataset for training and evaluation. The OUI dataset contains over 38k samples, representing a diverse demographic group in various environments. Models trained on this dataset have been reported to perform with accuracies of 80.6\% to 88.6\% depending on model architecture. After training the model structure described in Fig. \ref{fig:gender-classifier-structure} on the OUI dataset, the gender classification evaluated with accuracies of 0.848 and 0.807 on the DAiSEE training and validation sets, respectively. Because the model generalized on the DAiSEE dataset to within anticipated performance, and this performance is better than the initial gender classifier trained on DAiSEE, it is a more suitable gender classifier than a model trained on DAiSEE samples alone.

\begin{figure*}[t]
    \centering
    \includegraphics[scale=0.75]{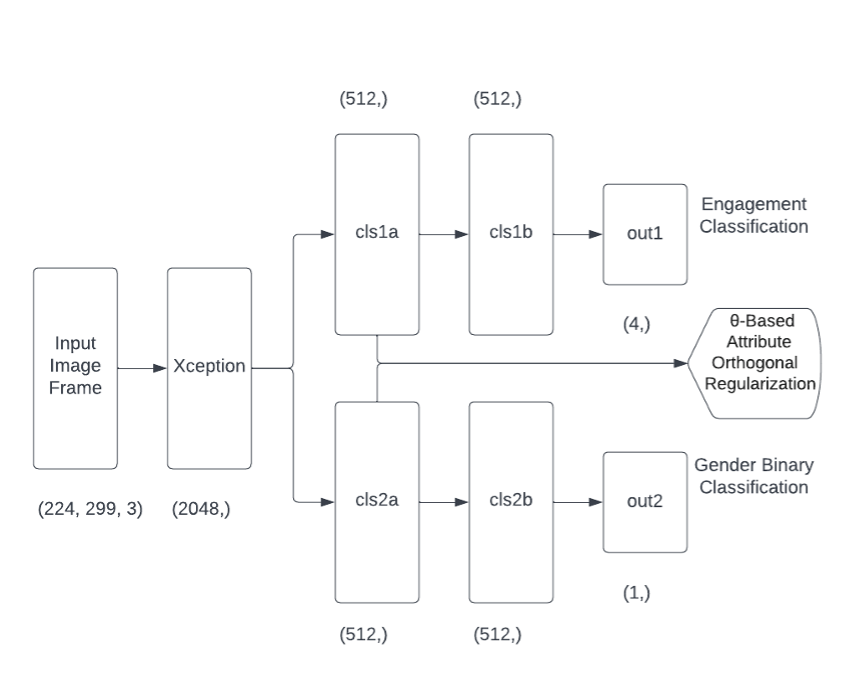}
    \caption{The proposed multi-task model}
    \label{fig:proposed-multi-task-model}
\end{figure*}

This additional step of training is beneficial to the concept of AOR mitigation. The first reason is because a gender classifier trained on the DAiSEE dataset is inherently subject to the skew in distribution of samples between genders. Secondly, the greater numbers of participants and scene conditions present in the external dataset creates a classifier that generalizes well. Additionally, the use of a separate dataset allows for future sensitivity groups to be included without adding additional labels to the DAiSEE dataset. This is a significant advantage, since the cost to add additional labels such as age, race, environment, etc. is not an obstacle in future bias mitigation tasks.
The layers of the gender classifier are then set to untrainable and new layers are added for engagement grading. The new layers are appended as a split-model branch, which bifurcates immediately after the last feature extractor layer (Fig. \ref{fig:proposed-multi-task-model}).

The regularization term is added to the categorical cross-entropy loss function used for engagement classification, and engagement training can then proceed. Mini-batch gradient descent was used to train the model, and throughout the training epochs, the loss function decreased on both training and validation set. Notably however, the accuracy on both training and validation datasets appeared to decrease slightly over time. This anomaly is discussed further in the next section.

\section{Results and Discussion}
\subsection{Distribution of Predictions}
The model trained with AOR is compared to a model trained without the regularization term. As a baseline model, we adopted earlier works \cite{gupta2016daisee,mohamad2020automatic,abedi2021improving} that combined a ResNet18 and Temporal Convolution Network (TCN), and the model shows an average accuracy, precision and recall on the four engagement levels of 55\%, 48\% and 38\%. We further grouped engagement levels 2 and 3 into a single group, and the recalculated accuracy got a bump to 93\%. However, we did not incorporate our AOR mitigation technique into the baseline model.

\begin{figure}
    \centering
    \includegraphics[scale=0.75]{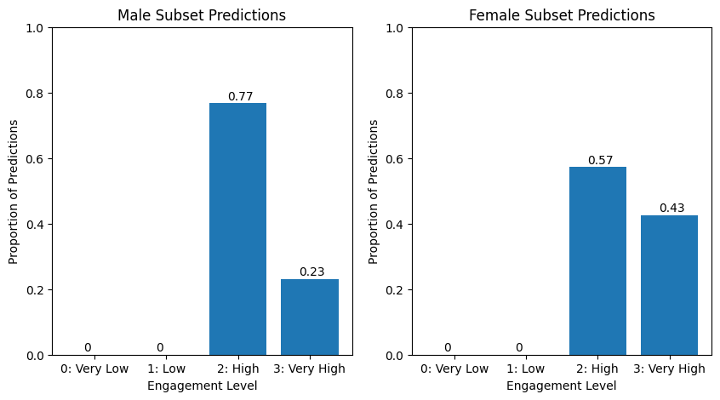}
    \caption{Non-mitigated Model Predictions}
    \label{fig:non-mitigated-model-preds}
\end{figure}

\begin{figure}
    \centering
    \includegraphics[scale=0.75]{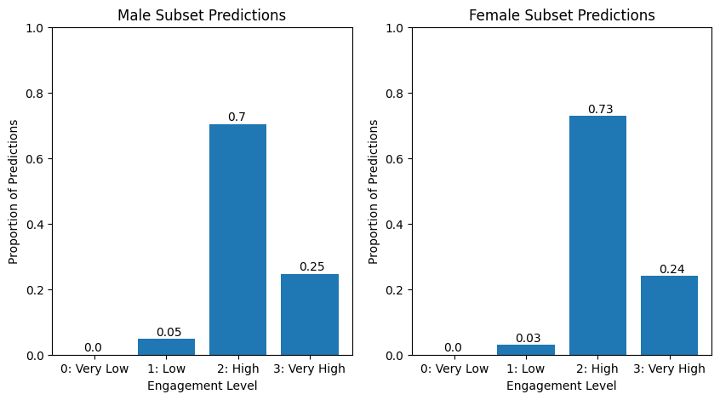}
    \caption{ AOR-mitigated Model Predictions}
    \label{fig:AOR-mitigated-model-preds}
\end{figure}

Fig. \ref{fig:non-mitigated-model-preds} and Fig. \ref{fig:AOR-mitigated-model-preds} list the normalized distribution of predictions on the validation set, broken down by gender. The histogram on the left shows the distribution of male predictions, while the histogram on the right shows the distribution of female predictions. 

To analyze the distribution of predictions between the sensitivity groups, the Pearson correlation coefficient (PCC) is used. This is performed by analyzing the two distributions as vectors. For non-mitigated male predictions, this vector is [0, 0, 0.77, 0.23] and for non-mitigated female predictions, this vector is [0, 0, 0.57, 0.43]. The PCC is 0.897 for the unmitigated model, and 0.999 for the mitigated model. This represents a significant improvement in reducing the difference in predictions between the sensitivity groups.

\subsection{$F_1$-Score Performance}
In order to gain additional insight, Fairlearn \cite{bird2020fairlearn} was used to evaluate the model. Surprisingly, according to the library, the disparity between predictions increased in some levels due to the use of AOR. Table \ref{tab:f1-comparison-mitigated-vs-unmitigated} illustrates $F_1$ scores per class as reported by the Fairlearn library.

\begin{table}[h]
    \centering
    \caption{$F_1$-score comparison between unmitigated and AOR-mitigated models}
    \begin{tabular}{ccccc}
    \toprule
       Set  &  Class 0 & Class 1 & Class 2 & Class 3  \\
    \midrule
    Unmitigated Male & 0 & 0 & 0.6797 & 0.2770\\
    Unmitigated Female & 0 & 0 & 0.4616 & 0.4998\\
    Mitigated Male & 0 & 0.0850 & 0.6600 & 0.2867\\
    Mitigated Female & 0 & 0.0276 & 0.4968 & 0.3681\\\bottomrule
    \end{tabular}
    
    \label{tab:f1-comparison-mitigated-vs-unmitigated}
\end{table}

The most significant impact in $F_1$-score is in the female very high engagement (level three) score. The drop in $F_1$-score from 0.499 to 0.3681 is significant, but explainable. Due to the skew that exists between gender labels in the validation set, this is to be expected. Because the ground truth for females was biased toward high engagement, and the model is performing predictions with gender bias mitigation practices in place, it is reasonable that the high engagement predictions would tend to be worse according to the validation set. To determine if this is a correct interpretation, future work will be needed to test out-of-distribution samples against the new model.

\subsection{Accuracy Performance}
Results regarding accuracy are inconclusive. Due to the skewed nature of the dataset, there are multiple factors which make interpreting the model performance based on an accuracy metric difficult.	

For a classification problem with four classes, and a dataset consisting of uniformly distributed labels, a random-guess prediction model would achieve an accuracy score of 25\%. For models trained on the DAiSEE dataset, a random-guess benchmark is complicated by the skew in the dataset. Levels two and three represent 94\% of the entire dataset, whereas level zero and one represent only 6\%. A model predicting only the mode of the distribution (level two) would achieve an accuracy score of 49\%. For comparison, the baseline model with unmitigated bias performed with an accuracy score of 54.9\%.

Through training with AOR, the model accuracy on the validation set achieved 54.5\% initially and dropped to 48\% through additional training. Because of the skew in distribution of ground true labels between the two sensitivity groups, it is uncertain whether this decrease in performance is representative only for in-dataset samples. It is possible that correcting for prediction bias inherent in the dataset is the cause for this decrease in performance on in-dataset samples. Further work is needed to validate how both models generalize for out-of-distribution samples.

\subsection{Uniform Distribution Performance}
\begin{figure}[t]
  \begin{subfigure}{.5\textwidth}
  \centering
    \includegraphics[width=1\linewidth]{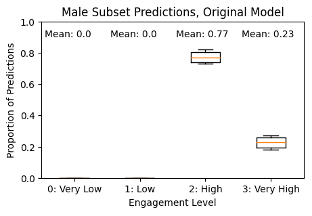}
  \end{subfigure}%
  \begin{subfigure}{.5\textwidth}
  \centering
    \includegraphics[width=1\linewidth]{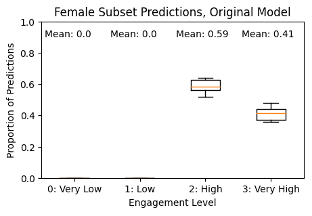}
  \end{subfigure}
  \caption{Predictions of unmitigated model by gender on a uniformly distributed dataset\label{fig:unmitigated-model-on-uniform-dataset}}
\end{figure}

\begin{figure}[t]
  \begin{subfigure}{.5\textwidth}
  \centering
    \includegraphics[width=1\linewidth]{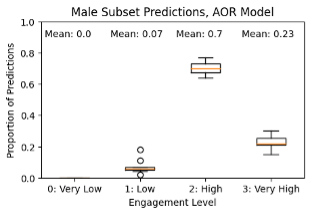}
  \end{subfigure}%
  \begin{subfigure}{.5\textwidth}
  \centering
    \includegraphics[width=1\linewidth]{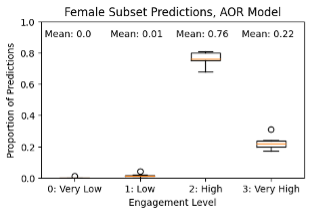}
  \end{subfigure}
  \caption{Predictions of AOR mitigated model by gender on a uniformly distributed dataset \label{fig:AOR-mitigated-model-on-uniform-dataset}}
\end{figure}

To reduce the impact of the label distribution skew on the prediction accuracy, a separate evaluation was performed on a uniformly distributed subset. The uniform subset is created by randomly selecting an equal number of samples from each intersection of engagement level and gender. For example, this new subset contains the same number of very low male examples as the number of very low female examples, as well as an equal number between engagement levels. Because there are fewer samples present in some intersections, this limited the overall number of samples that could be used in the uniform subset. The total number of samples in this subset is 168. This constrained subset is drawn randomly from the validation set and evaluated 10 times to average the results. The proportions of predictions on this subset are similar to predictions made on the overall validation set. Fig. \ref{fig:unmitigated-model-on-uniform-dataset} shows the proportions of predictions of the unmitigated model for each engagement level, broken down by gender. Similarly, Fig. \ref{fig:AOR-mitigated-model-on-uniform-dataset} shows the proportions of predictions of the AOR mitigated model for each engagement level, broken down by gender. Even though these predictions are performed on a uniformly distributed dataset, there is a clear bias demonstrated in the original, unmitigated model that predicts males to be, on average, less engaged than females. In the AOR mitigated model, the predictions were more similar between genders. This indicates that the bias mitigation technique was successful in reducing the bias in predictions. 

\section{Conclusions and Future works} 
Exploratory data analysis of the DAiSEE dataset has yielded information pertaining to the present skew in distribution. Additionally, Grad-CAM analysis of a DNN model trained on the dataset has shown that the model performs differently between male and female samples. This issue is highlighted in the distribution of predictions. Additionally, issues such as the temporal nature of the samples is considered, and solutions are proposed.

Models trained with AOR mitigation techniques show decreased prediction bias. Using a multi-task, split-model training regimen has been shown to improve the results of models trained with AOR. This technique is suitable for mitigating bias in vision tasks and is potentially extensible to a variety of sensitivity groups. 

Additional research is needed to model the AOR hyperparameter, $\lambda$ in the objective function which scales the amount of influence the $L_\text{ortho}$ term effects the loss. This will require retraining multiple models at varying $\alpha$ from zero to some arbitrary value.  Additionally, to investigate how AOR performs with other datasets and tasks, the technique should be implemented on a variety of models across multiple datasets to see how the technique generalizes across different tasks. 

Furthermore, there would be value in using AOR with models outside of visual models with simple MLPs. 
The effect of fine-tuning on CNN layers for either or both classifiers could yield significant improvements. There is evidence that specialization occurs in particular neurons or filters in DNNs, similar to the concept in neuroscience of “grand-mother cells” (GMC) \cite{le2013building}. This term derives its name from the concept that there are cells in the human brain that only respond to a specific stimulus, such as the face of a person’s grandmother. It is possible that neurons or filters trained specifically for the sensitivity group classifier could improve the performance of models using AOR.

As mentioned in section 3, additional research is needed for the DAiSEE dataset alone. A validation study would be helpful in building models based on the dataset. A small subsample of the of dataset should be examined by a group of study participants to verify the ground truth labels of the samples. This would provide a baseline for a DNN model prediction metric for comparison. Additionally, it would help validate whether the bias introduced is through label generation is a product of cultural bias, or if the ground truths are accurate.


\section*{ACKNOWLEDGMENT}
This material is based upon work supported by the National Science Foundation under Grant No. 2329919.

\bibliographystyle{plainnat}
\bibliography{references}

\begin{thebibliography}{23}
\providecommand{\natexlab}[1]{#1}
\providecommand{\url}[1]{\texttt{#1}}
\expandafter\ifx\csname urlstyle\endcsname\relax
  \providecommand{\doi}[1]{doi: #1}\else
  \providecommand{\doi}{doi: \begingroup \urlstyle{rm}\Url}\fi

\bibitem[Abedi and Khan(2021)]{abedi2021improving}
Ali Abedi and Shehroz~S Khan.
\newblock Improving state-of-the-art in detecting student engagement with
  resnet and tcn hybrid network.
\newblock In \emph{2021 18th Conference on Robots and Vision (CRV)}, pages
  151--157. IEEE, 2021.

\bibitem[Agrawal et~al.(2014)Agrawal, Girshick, and
  Malik]{agrawal2014analyzing}
Pulkit Agrawal, Ross Girshick, and Jitendra Malik.
\newblock Analyzing the performance of multilayer neural networks for object
  recognition.
\newblock In \emph{Computer Vision--ECCV 2014: 13th European Conference,
  Zurich, Switzerland, September 6-12, 2014, Proceedings, Part VII 13}, pages
  329--344. Springer, 2014.

\bibitem[Akhand et~al.(2021)Akhand, Roy, Siddique, Kamal, and
  Shimamura]{akhand2021facial}
MAH Akhand, Shuvendu Roy, Nazmul Siddique, Md~Abdus~Samad Kamal, and Tetsuya
  Shimamura.
\newblock Facial emotion recognition using transfer learning in the deep cnn.
\newblock \emph{Electronics}, 10\penalty0 (9):\penalty0 1036, 2021.

\bibitem[Arjovsky et~al.(2019)Arjovsky, Bottou, Gulrajani, and
  Lopez-Paz]{arjovsky2019invariant}
Martin Arjovsky, L{\'e}on Bottou, Ishaan Gulrajani, and David Lopez-Paz.
\newblock Invariant risk minimization.
\newblock \emph{arXiv preprint arXiv:1907.02893}, 2019.

\bibitem[Bird et~al.(2020)Bird, Dud{\'\i}k, Edgar, Horn, Lutz, Milan, Sameki,
  Wallach, and Walker]{bird2020fairlearn}
Sarah Bird, Miro Dud{\'\i}k, Richard Edgar, Brandon Horn, Roman Lutz, Vanessa
  Milan, Mehrnoosh Sameki, Hanna Wallach, and Kathleen Walker.
\newblock Fairlearn: A toolkit for assessing and improving fairness in ai.
\newblock \emph{Microsoft, Tech. Rep. MSR-TR-2020-32}, 2020.

\bibitem[Deng et~al.(2009)Deng, Dong, Socher, Li, Li, and
  Fei-Fei]{deng2009imagenet}
Jia Deng, Wei Dong, Richard Socher, Li-Jia Li, Kai Li, and Li~Fei-Fei.
\newblock Imagenet: A large-scale hierarchical image database.
\newblock In \emph{2009 IEEE conference on computer vision and pattern
  recognition}, pages 248--255. Ieee, 2009.

\bibitem[Eidinger et~al.(2014)Eidinger, Enbar, and Hassner]{eidinger2014age}
Eran Eidinger, Roee Enbar, and Tal Hassner.
\newblock Age and gender estimation of unfiltered faces.
\newblock \emph{IEEE Transactions on information forensics and security},
  9\penalty0 (12):\penalty0 2170--2179, 2014.

\bibitem[Fran et~al.(2017)]{fran2017deep}
C~Fran et~al.
\newblock Deep learning with depth wise separable convolutions.
\newblock In \emph{IEEE conference on computer vision and pattern recognition
  (CVPR)}, 2017.

\bibitem[Goeleven et~al.(2008)Goeleven, De~Raedt, Leyman, and
  Verschuere]{goeleven2008karolinska}
Ellen Goeleven, Rudi De~Raedt, Lemke Leyman, and Bruno Verschuere.
\newblock The karolinska directed emotional faces: a validation study.
\newblock \emph{Cognition and emotion}, 22\penalty0 (6):\penalty0 1094--1118,
  2008.

\bibitem[Gupta et~al.(2016)Gupta, Jaiswal, Adhikari, and
  Balasubramanian]{gupta2016daisee}
Abhay Gupta, Richik Jaiswal, Sagar Adhikari, and Vineeth Balasubramanian.
\newblock Daisee: dataset for affective states in e-learning environments.
\newblock \emph{arXiv preprint arXiv:1609.01885}, pages 1--22, 2016.

\bibitem[Haig(2003)]{haig2003spurious}
Brian~D Haig.
\newblock What is a spurious correlation?
\newblock \emph{Understanding Statistics: Statistical Issues in Psychology,
  Education, and the Social Sciences}, 2\penalty0 (2):\penalty0 125--132, 2003.

\bibitem[Khani and Liang(2021)]{khani2021removing}
Fereshte Khani and Percy Liang.
\newblock Removing spurious features can hurt accuracy and affect groups
  disproportionately.
\newblock In \emph{Proceedings of the 2021 ACM conference on fairness,
  accountability, and transparency}, pages 196--205, 2021.

\bibitem[Le(2013)]{le2013building}
Quoc~V Le.
\newblock Building high-level features using large scale unsupervised learning.
\newblock In \emph{2013 IEEE international conference on acoustics, speech and
  signal processing}, pages 8595--8598. IEEE, 2013.

\bibitem[Lou et~al.(2013)Lou, Caruana, Gehrke, and Hooker]{lou2013accurate}
Yin Lou, Rich Caruana, Johannes Gehrke, and Giles Hooker.
\newblock Accurate intelligible models with pairwise interactions.
\newblock In \emph{Proceedings of the 19th ACM SIGKDD international conference
  on Knowledge discovery and data mining}, pages 623--631, 2013.

\bibitem[Lundqvist et~al.(1998)Lundqvist, Flykt, and
  {\"O}hman]{lundqvist1998karolinska}
Daniel Lundqvist, Anders Flykt, and Arne {\"O}hman.
\newblock Karolinska directed emotional faces.
\newblock \emph{PsycTESTS Dataset}, 91:\penalty0 630, 1998.

\bibitem[Madan et~al.(2020)Madan, Henry, Dozier, Ho, Bhandari, Sasaki, Durand,
  Pfister, and Boix]{madan2020and}
Spandan Madan, Timothy Henry, Jamell Dozier, Helen Ho, Nishchal Bhandari,
  Tomotake Sasaki, Fr{\'e}do Durand, Hanspeter Pfister, and Xavier Boix.
\newblock When and how cnns generalize to out-of-distribution
  category-viewpoint combinations.
\newblock \emph{arXiv preprint arXiv:2007.08032}, 2020.

\bibitem[Mohamad~Nezami et~al.(2020)Mohamad~Nezami, Dras, Hamey, Richards, Wan,
  and Paris]{mohamad2020automatic}
Omid Mohamad~Nezami, Mark Dras, Len Hamey, Deborah Richards, Stephen Wan, and
  C{\'e}cile Paris.
\newblock Automatic recognition of student engagement using deep learning and
  facial expression.
\newblock In \emph{Joint european conference on machine learning and knowledge
  discovery in databases}, pages 273--289. Springer, 2020.

\bibitem[Pan and Yang(2009)]{pan2009survey}
Sinno~Jialin Pan and Qiang Yang.
\newblock A survey on transfer learning.
\newblock \emph{IEEE Transactions on knowledge and data engineering},
  22\penalty0 (10):\penalty0 1345--1359, 2009.

\bibitem[Selvaraju et~al.(2017)Selvaraju, Cogswell, Das, Vedantam, Parikh, and
  Batra]{selvaraju2017grad}
Ramprasaath~R Selvaraju, Michael Cogswell, Abhishek Das, Ramakrishna Vedantam,
  Devi Parikh, and Dhruv Batra.
\newblock Grad-cam: Visual explanations from deep networks via gradient-based
  localization.
\newblock In \emph{Proceedings of the IEEE international conference on computer
  vision}, pages 618--626, 2017.

\bibitem[Singla and Feizi(2021)]{singla2021salient}
Sahil Singla and Soheil Feizi.
\newblock Salient imagenet: How to discover spurious features in deep learning?
\newblock \emph{arXiv preprint arXiv:2110.04301}, 2021.

\bibitem[Teotia et~al.(2022)Teotia, Mao, and Vondrick]{teotia2022finding}
Revant Teotia, Chengzhi Mao, and Carl Vondrick.
\newblock Finding spuriously correlated visual attributes.
\newblock In \emph{ICML 2022: Workshop on Spurious Correlations, Invariance and
  Stability}, 2022.

\bibitem[Wang et~al.(2019)Wang, Zhao, Yatskar, Chang, and
  Ordonez]{wang2019balanced}
Tianlu Wang, Jieyu Zhao, Mark Yatskar, Kai-Wei Chang, and Vicente Ordonez.
\newblock Balanced datasets are not enough: Estimating and mitigating gender
  bias in deep image representations.
\newblock In \emph{Proceedings of the IEEE/CVF international conference on
  computer vision}, pages 5310--5319, 2019.

\bibitem[Yang et~al.(2022)Yang, Gupta, Feng, Singhal, Yadav, Wu, Natarajan,
  Hedau, and Joo]{yang2022enhancing}
Yu~Yang, Aayush Gupta, Jianwei Feng, Prateek Singhal, Vivek Yadav, Yue Wu,
  Pradeep Natarajan, Varsha Hedau, and Jungseock Joo.
\newblock Enhancing fairness in face detection in computer vision systems by
  demographic bias mitigation.
\newblock In \emph{Proceedings of the 2022 AAAI/ACM Conference on AI, Ethics,
  and Society}, pages 813--822, 2022.

\end{thebibliography}

\end{document}